\PassOptionsToPackage{table}{xcolor}

\documentclass[letterpaper, 10 pt, conference]{class/ieeeconf}  

\IEEEoverridecommandlockouts                              
\usepackage{color}
\usepackage[table,xcdraw]{xcolor}
\usepackage{amsmath}
\usepackage{amssymb}
\usepackage{latexsym}
\usepackage{url}
\usepackage{cite}
\usepackage{relsize}
\usepackage{multirow}
\usepackage{afterpage}
\usepackage{ifthen}
\usepackage{graphicx}
\usepackage{algpseudocode,algorithm,algorithmicx}
\usepackage{subfigure}
\usepackage{epstopdf}
\usepackage{stackengine}
\usepackage{gensymb}
\usepackage{booktabs}
\usepackage{makecell}
\usepackage{hhline}
\usepackage{lipsum}
\usepackage{ upgreek }
\usepackage{mathtools}
\newcommand{\etal}{\textit{et al.}}
\newcommand{\improvecolor}{\color[HTML]{3B9612}}

\makeatletter
\let\NAT@parse\undefined
\makeatother
\usepackage[bookmarks=false, linkcolor=blue, urlcolor=blue, citecolor=blue]{hyperref} 

\hypersetup{
    colorlinks=true,
    linkcolor=red,
    filecolor=magenta,      
    urlcolor=blue,
    pdfstartview={FitH},
    citecolor =blue
    }

\title{\LARGE \bf Reducing Non-IID Effects in Federated Autonomous Driving with Contrastive Divergence Loss}

\author{Tuong Do$^{1,3}$, Binh X. Nguyen$^{1}$, Quang D. Tran$^{1}$, Hien Nguyen$^{1}$, Erman Tjiputra$^{1}$, \\Te-Chuan Chiu$^{2}$, Anh Nguyen$^{3}$
\thanks{$^{1}$AIOZ, Singapore 
        {\tt\small tuong.khanh-long.do@aioz.io}}%
\thanks{$^{2}$Department of Computer Science, National Tsing Hua University, Taiwan
        {\tt\small theochiu@cs.nthu.edu.tw}}%
\thanks{$^{3}$Deparment of Computer Science, University of Liverpool, UK
        {\tt\small anh.nguyen@liverpool.ac.uk}
        }%
}

\begin{document}

\newtheorem{problem}{Problem}
\newtheorem{lemma}{Lemma}
\newtheorem{theorem}[lemma]{Theorem}
\newtheorem{claim}{Claim}
\newtheorem{corollary}[lemma]{Corollary}
\newtheorem{definition}[lemma]{Definition}
\newtheorem{proposition}[lemma]{Proposition}
\newtheorem{remark}[lemma]{Remark}
\newenvironment{LabeledProof}[1]{\noindent{\it Proof of #1: }}{\qed}

\def\beq#1\eeq{\begin{equation}#1\end{equation}}
\def\bea#1\eea{\begin{align}#1\end{align}}
\def\beg#1\eeg{\begin{gather}#1\end{gather}}
\def\beqs#1\eeqs{\begin{equation*}#1\end{equation*}}
\def\beas#1\eeas{\begin{align*}#1\end{align*}}
\def\begs#1\eegs{\begin{gather*}#1\end{gather*}}

\newcommand{\poly}{\mathrm{poly}}
\newcommand{\eps}{\epsilon}
\newcommand{\e}{\epsilon}
\newcommand{\polylog}{\mathrm{polylog}}
\newcommand{\rob}[1]{\left( #1 \right)} 
\newcommand{\sqb}[1]{\left[ #1 \right]} 
\newcommand{\cub}[1]{\left\{ #1 \right\} } 
\newcommand{\rb}[1]{\left( #1 \right)} 
\newcommand{\abs}[1]{\left| #1 \right|} 
\newcommand{\zo}{\{0, 1\}}
\newcommand{\zonzo}{\zo^n \to \zo}
\newcommand{\zokzo}{\zo^k \to \zo}
\newcommand{\zot}{\{0,1,2\}}
\newcommand{\en}[1]{\marginpar{\textbf{#1}}}
\newcommand{\efn}[1]{\footnote{\textbf{#1}}}
\newcommand{\vecbm}[1]{\boldmath{#1}} 
\newcommand{\uvec}[1]{\hat{\vec{#1}}}
\newcommand{\thv}{\vecbm{\theta}}
\newcommand{\junk}[1]{}
\newcommand{\var}{\mathop{\mathrm{var}}}
\newcommand{\rank}{\mathop{\mathrm{rank}}}
\newcommand{\diag}{\mathop{\mathrm{diag}}}
\newcommand{\tr}{\mathop{\mathrm{tr}}}
\newcommand{\acos}{\mathop{\mathrm{acos}}}
\newcommand{\atantwo}{\mathop{\mathrm{atan2}}}
\newcommand{\SVD}{\mathop{\mathrm{SVD}}}
\newcommand{\quadf}{\mathop{\mathrm{q}}}
\newcommand{\linterp}{\mathop{\mathrm{l}}}
\newcommand{\sgn}{\mathop{\mathrm{sign}}}
\newcommand{\sym}{\mathop{\mathrm{sym}}}
\newcommand{\avg}{\mathop{\mathrm{avg}}}
\newcommand{\mean}{\mathop{\mathrm{mean}}}
\newcommand{\erf}{\mathop{\mathrm{erf}}}
\newcommand{\grad}{\nabla}
\newcommand{\R}{\mathbb{R}}
\newcommand{\defeq}{\triangleq}
\newcommand{\dims}[2]{[#1\!\times\!#2]}
\newcommand{\sdims}[2]{\mathsmaller{#1\!\times\!#2}}
\newcommand{\udims}[3]{#1}
\newcommand{\udimst}[4]{#1}
\newcommand{\com}[1]{\rhd\text{\emph{#1}}}
\newcommand{\ind}{\hspace{1em}}
\newcommand{\argmin}[1]{\underset{#1}{\operatorname{argmin}}}
\newcommand{\floor}[1]{\left\lfloor{#1}\right\rfloor}
\newcommand{\step}[1]{\vspace{0.5em}\noindent{#1}}
\newcommand{\quat}[1]{\ensuremath{\mathring{\mathbf{#1}}}}
\newcommand{\norm}[1]{\left\lVert#1\right\rVert}
\newcommand{\ignore}[1]{}
\newcommand{\specialcell}[2][c]{\begin{tabular}[#1]{@{}c@{}}#2\end{tabular}}
\newcommand*\Let[2]{\State #1 $\gets$ #2}
\newcommand{\algorithmicbreak}{\textbf{break}}
\newcommand{\Break}{\State \algorithmicbreak}
\newcommand{\ra}[1]{\renewcommand{\arraystretch}{#1}}

\renewcommand{\vec}[1]{\mathbf{#1}} 

\algdef{S}[FOR]{ForEach}[1]{\algorithmicforeach\ #1\ \algorithmicdo}
\algnewcommand\algorithmicforeach{\textbf{for each}}
\algrenewcommand\algorithmicrequire{\textbf{Require:}}
\algrenewcommand\algorithmicensure{\textbf{Ensure:}}
\algnewcommand\algorithmicinput{\textbf{Input:}}
\algnewcommand\INPUT{\item[\algorithmicinput]}
\algnewcommand\algorithmicoutput{\textbf{Output:}}
\algnewcommand\OUTPUT{\item[\algorithmicoutput]}

\maketitle
\thispagestyle{empty}
\pagestyle{empty}

\begin{abstract}
Federated learning has been widely applied in autonomous driving since it enables training a learning model among vehicles without sharing users' data. However, data from autonomous vehicles usually suffer from the non-independent-and-identically-distributed (non-IID) problem, which may cause negative effects on the convergence of the learning process. In this paper, we propose a new contrastive divergence loss to address the non-IID problem in autonomous driving by reducing the impact of divergence factors from transmitted models during the local learning process of each silo. We also analyze the effects of contrastive divergence in various autonomous driving scenarios, under multiple network infrastructures, and with different centralized/distributed learning schemes. Our intensive experiments on three datasets demonstrate that our proposed contrastive divergence loss significantly improves the performance over current state-of-the-art approaches.
\end{abstract}

\section{Introduction}
Autonomous driving is an emerging field that enables vehicles to operate without a human driver by using a combination of vision, learning, and control algorithms to observe and respond to changes in the environment~\cite{grigorescu2020survey}. Recently, many works have been proposed to address different problems in autonomous driving~\cite{bergamini2021simnet,chen2022pseudo,wang2022ltp}. While significant progress has been made in the field, traditional works utilize supervised learning methods and require data collection to train the model~\cite{huang2021joint,vitelli2022safetynet}. Although collecting data is necessary to improve the system's accuracy, it strongly violates user privacy since the users' data are shared with third parties. Recent works have adapted federated learning (FL) as a new learning mechanism to overcome this limitation. Federated learning allows multiple parties to collaboratively train a model without sharing their data~\cite{nguyen2022deep,liang2019federated,li2021privacy,nguyen2022multigraph}. In practice, federated learning enables autonomous vehicles to learn a shared prediction model together, involving more diverse data while preserving users' privacy~\cite{zhang2021survey}.

\begin{figure}[t]
  \centering
\resizebox{\linewidth}{!}{
\setlength{\tabcolsep}{0.5pt}
\begin{tabular}{ccc}
\shortstack{\includegraphics[width=0.33\linewidth]{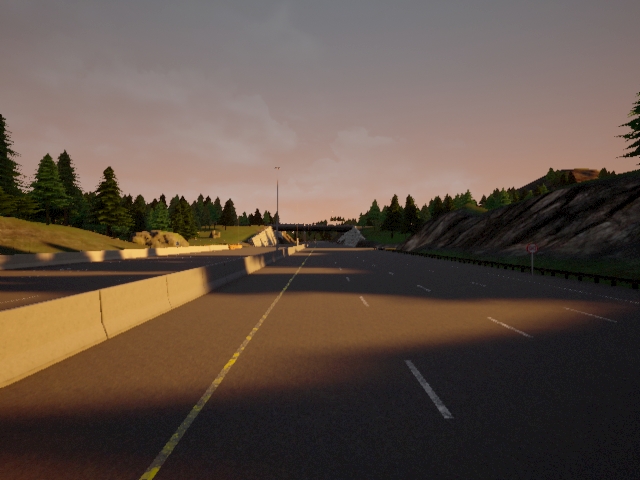}}&
\shortstack{\includegraphics[width=0.33\linewidth]{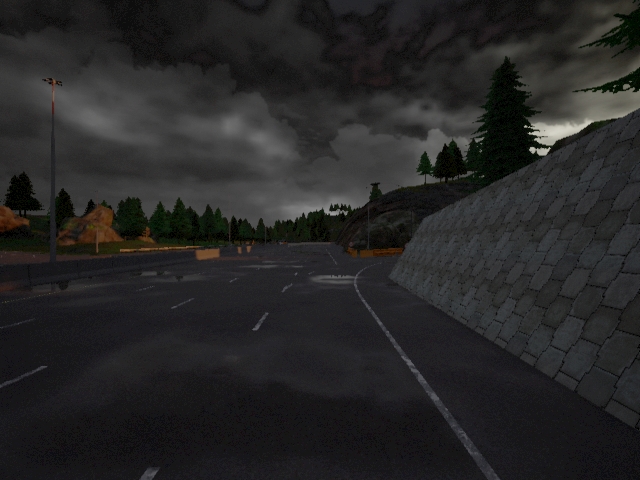}}&
\shortstack{\includegraphics[width=0.33\linewidth]{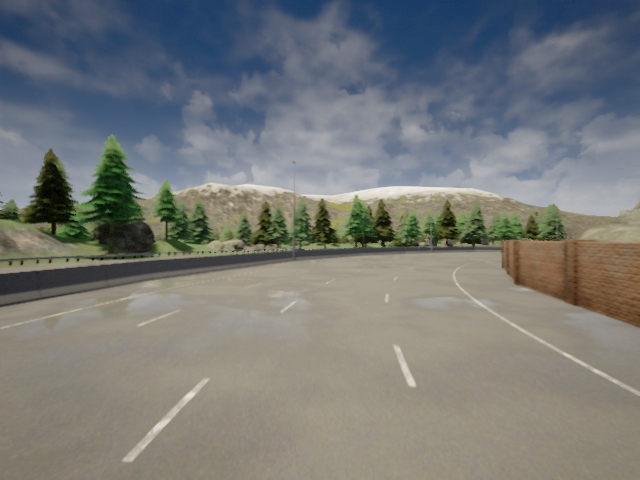}}\\
\shortstack{\includegraphics[width=0.33\linewidth]{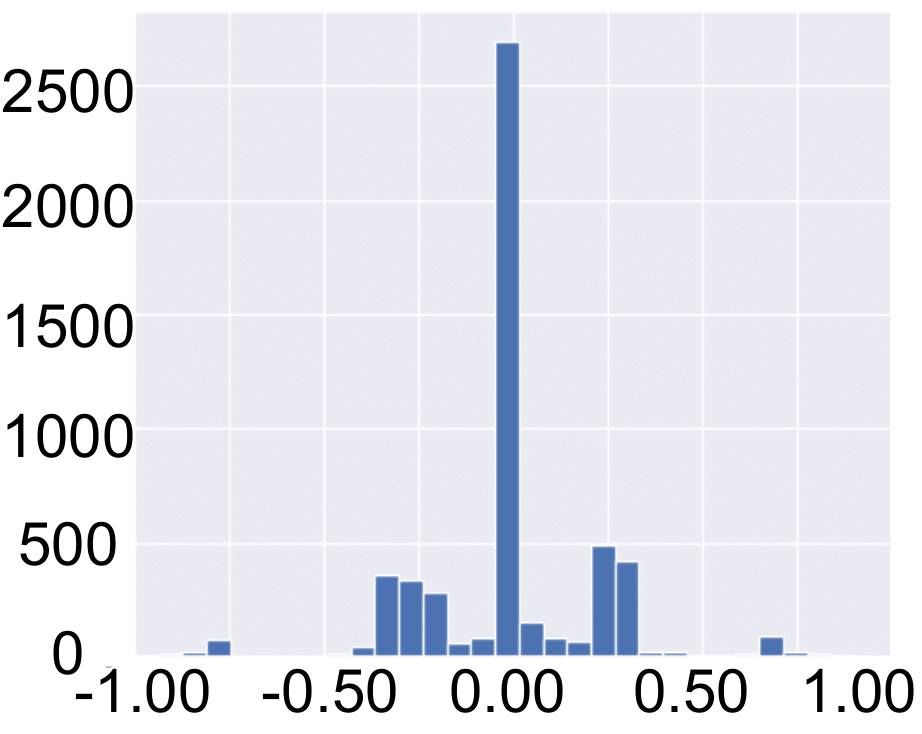}}&
\shortstack{\includegraphics[width=0.33\linewidth]{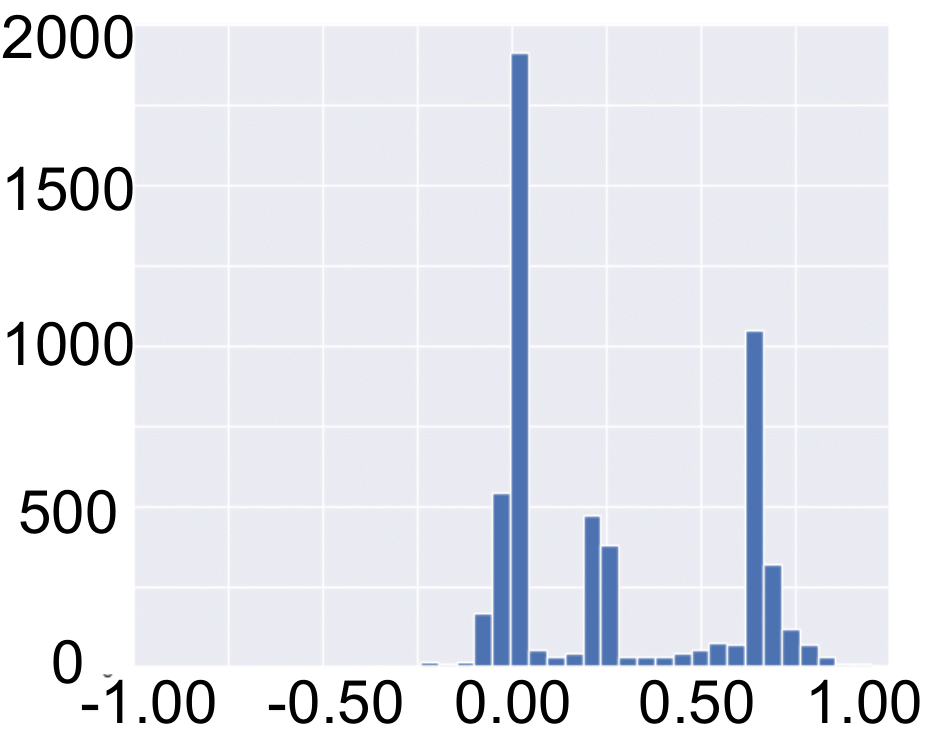}}&
\shortstack{\includegraphics[width=0.33\linewidth]{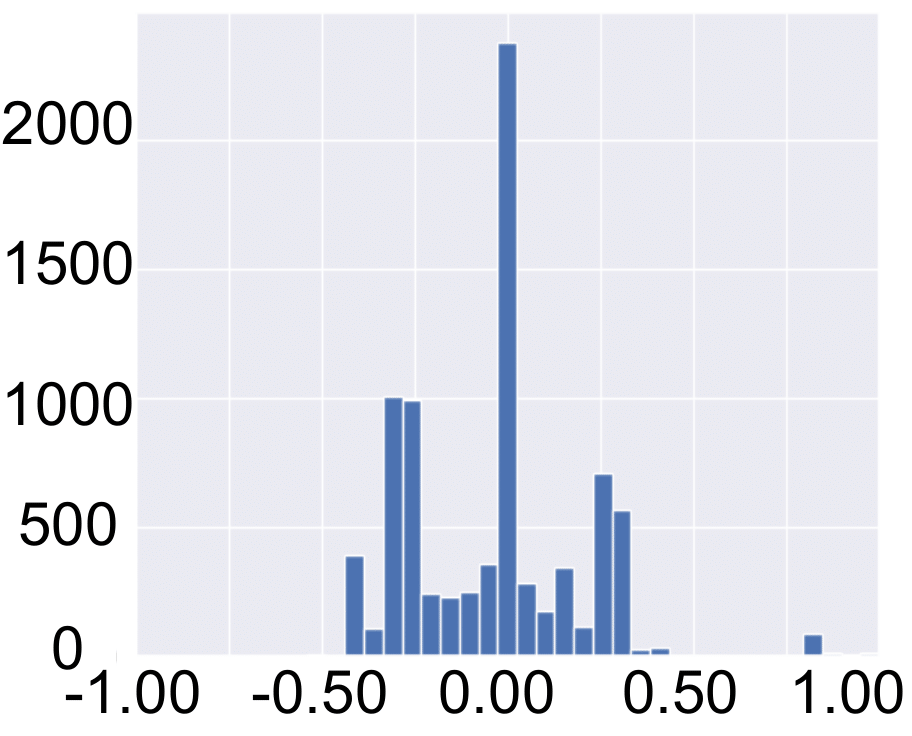}}\\
\end{tabular}
}
\vspace{-2ex}
    \caption{Sample viewpoints over three vehicles and their steering angle distributions in Carla dataset~\cite{nguyen2022deep}. The visualization shows that the three vehicles have differences in visual input as well as steering angle distribution.}
    \label{fig:intro_fig}
\end{figure}

 
Typically, there are two main federated learning scenarios~\cite{marfoq2020throughput}: Server-based Federated Learning (SFL) which has a central node to orchestrate the training process and receive the contributions of all clients, and Decentralized Federated Learning (DFL) which utilizes a fully peer-to-peer (P2P) setup between data silos using a predefined topology. Although SFL can enhance data privacy since only model weights are transmitted, having orchestration nodes potentially represents a bottleneck of the system since most of the data transmission has to go through the central nodes. On the other hand, DFL does not require a server and uses a fully distributed network. 
In autonomous driving, several works have explored both DFL and SFL to address different problems such as collision avoidance~\cite{liang2019federated,hammedi2022toward}, trajectory prediction~\cite{yin2022trajectory}, and steering prediction~\cite{li2021privacy,nguyen2022deep}.


In practice, while SFL or DFL approaches have their own advantages and limitations, both of them suffer from the non-IID problem in federated learning. According to~\cite{sattler2019robust,wang2018cooperative}, the \textit{non-IID} (identically and independently distributed) problem occurs when data partitioning across silos has a significant \textit{distribution shift}. Although the non-IID problem occurred in many contexts, it is an immense problem in autonomous driving and causes difficulties when the accumulation process for all vehicle silos is conducted~\cite{li2022feel}. In practice, each vehicle has its unique driving patterns, weather conditions, and road types, which can cause differences in the data distribution. For example, data collected from a car driving on a highway may be different from data collected from a car driving in a city center. When the accumulation process for all vehicles is conducted, the non-IID problem can cause difficulties in building robust and accurate learning models, e.g., if the model is trained mainly on data from highways, it may perform poorly in urban environments. Figure~\ref{fig:intro_fig} illustrates different scenarios when we have the non-IID problem in autonomous driving.



Prior works that address the non-IID problem by optimizing the accumulation step~\cite{zhang2021end,li2022feel,mu2023fedproc}, adding normalization to the global model~\cite{idrissi2021fedbs}, fine-tuning global model weights through distillation~\cite{zhang2022fine}, or aligning the distribution between the global model and local ones~\cite{gao2022feddc,zhang2021federated}. In general, these solutions try to adapt to divergence factors by refining the global model weights. However, this process can be intricate for optimization since the global model receives weights learned from diverse distributions of distinct local datasets~\cite{zhu2021federated}. Consequently, these methods encounter unresolved hurdles such as identifying the ideal accumulation step size tailored to each local silo, dealing with disconnection and bottleneck in the network topology, or ensuring the consistency between local and global objectives.

In this paper, we propose a new Contrastive Divergence Loss (CDL) to address the non-IID problem in autonomous driving. Instead of \textit{waiting for the mode}l to be learned locally and then dealing with the Non-IID problem during the global aggregating process as other methods~\cite{li2022feel,tan2022federated,you2022reschedule}, we \textit{directly mitigate the impact of divergence factors} throughout the learning process of each local silo using our proposed CDL loss. Compared with dealing Non-IID problem in the global model~\cite{yang2021achieving,tan2022federated,you2022reschedule,mu2023fedproc}, facilitating the adaptation of distribution between neighboring silos is less complex. In this way, each local model is more robust to changes in the data distribution and we can use typical accumulation methods (e.g., FedAvg~\cite{li2019convergence}) to conduct global weights without concerning the Non-IID effects on the convergence. 
The intensive experiments on three autonomous driving datasets verify our observation and show significant improvements over state-of-the-art methods.

\section{Related works}
\textbf{Autonomous Driving.}
Autonomous driving is an emerging field that has attracted significant research interests in recent years. Several works have focused on using deep learning for autonomous vehicle control, such as 
object detection and tracking~\cite{hu2022investigating,chen2022pseudo}, trajectory prediction~\cite{wang2022ltp}, autonomous braking and steering~\cite{ijaz2021automatic}. Recently, Xin \etal~\cite{xin2020slip} proposed a recursive backstepping steering controller that effectively links yaw-rate-based path following commands to the steering angle. Xiong \etal~\cite{xiong2021reduced} analyzed the nonlinear dynamics behavior using a proportional control law. Yi \etal~\cite{yi2022anti} presented an algorithm to select the instantaneous center of rotation within the self-reconfigurable robot's area and perform static rotation to adjust its heading angle during waypoint navigation while avoiding collisions. Recently, Yin \etal~\cite{yin2022trajectory} combined model predictive control with covariance steering theory to obtain a robust controller for general nonlinear autonomous driving systems.

\textbf{Federated Learning for Autonomous Driving.} 
Federated learning allows many participants to train a machine learning model cooperatively without disclosing their local data~\cite{zhang2021survey}.
Federated learning offers a privacy-aware solution to many automotive systems, such as cooperative autonomous driving and intelligent transport systems, which require efficient communication, computation, and storage~\cite{du2020federated}. Recently, many works have been proposed to address different problems in autonomous driving using federated learning~\cite{zhang2021real,he2021bift,khan2022journey,liang2022federated,parekh2023gefl}. Liang \etal~\cite{liang2019federated} presented an online federated reinforcement transfer learning process where all the vehicles make actions from the knowledge learned by others. The authors in~\cite{li2021privacy} proposed an autonomous driving system to preserve vehicle privacy by using a server to store shared training models between vehicles. Zhang \etal~\cite{zhang2021real} proposed an approach that included a unique asynchronous model aggregation mechanism. The authors in~\cite{nguyen2022deep} introduced a deep federated network for steering angle prediction. Recently, Doomra~\etal~\cite{doomra2020turn} used federated learning to predict the turning signal. 

Although applying federated learning in autonomous driving is trendy, there are several open challenges. Besides the crucial disadvantages correlated to the availability and quality of the local computing devices, the variability in data distribution across the participants is also challenging. Specifically, the data of various vehicles are usually non-IID. While non-IID is a critical problem, most of the recent works focus on addressing it through the accumulation process~\cite{zhang2021end,li2022feel,li2022enhancing,mu2023fedproc} and do not focus on local silos optimization.

\textbf{Contrastive Divergence.} 
Contrastive models have been studied in FL recently for dealing with the heterogeneity of local data distribution across parties~\cite{li2021privacy}. This type of model has been applied to vision datasets~\cite{mu2023fedproc}, language datasets~\cite{georgiev2013non}, and signal datasets~\cite{tsouvalas2022federated}. Indeed, contrastive learning is a valuable candidate for dealing with the non-IID problem in federated autonomous driving, which is mostly caused by local data heterogeneity. It is worth noting that most contrastive-related works focus on building optimized frameworks in server-based~\cite{tan2022federated, qifairvfl}, modifying topology in P2P federated learning~\cite{bellet2021d,mu2023fedproc}, or adjusting the accumulation process~\cite{gao2022feddc,kassem2022federated, hu2023fedssc}. However,  as far as our knowledge, no works have considered the contrastive behavior of the loss function to reduce the effect of the Non-IID data in the federated autonomous driving scenario.


\section{Preliminary}
\label{sec:preliminary}

\subsection{Notation} 
We summarize the notations of our paper in Table~\ref{tab:notation}. 

\begin{table}[h]
\caption{Mathematical Notations.}
\begin{center}
\setlength{\tabcolsep}{0.3 em} 
{\renewcommand{\arraystretch}{1.0}
\begin{tabular}{cl|cl}
\toprule 
Not. & Description & Not. & Description \cr\hline
\midrule 
$\xi$    &\begin{tabular}[c]{@{}l@{}}Data in a mini-batch \end{tabular}                                            &$i,j$       & Workers (silos) \\

\rowcolor[HTML]{EFEFEF}$\alpha$                & Learning rate                                    & $m$         &Mini-batch size  \\
$\textbf{A}$               & Consensus matrix                                  & $\mathcal{L}$         &Loss \\
\rowcolor[HTML]{EFEFEF}  $k$            &\begin{tabular}[c]{@{}l@{}} One specific iteration\end{tabular}                                & $\vartheta \in \{0,1\}$    &\begin{tabular}[c]{@{}l@{}}Aggregation status \end{tabular}  \\
$x$    &\begin{tabular}[c]{@{}l@{}}Input image of $\xi$ \\in $k$-th iteration\end{tabular}                                            &$x'$       &\begin{tabular}[c]{@{}l@{}}Input image of $\xi$\\ in $(k-1)$-th iteration \end{tabular}   \\
\rowcolor[HTML]{EFEFEF}$\mathcal{H}$          &\begin{tabular}[c]{@{}l@{}} Kullback-Leibler\\ Divergence\end{tabular}                                &$\theta$                  &\begin{tabular}[c]{@{}l@{}} Learnable\\ parameters\end{tabular}    \\
$\beta$           &\begin{tabular}[c]{@{}l@{}} Pulling control parameter for\\  Kullback-Leibler Divergence\end{tabular}                                & $\mathcal{N}_i^{+}$                     &\begin{tabular}[c]{@{}l@{}} In-neighbors \\ of silo $i$ \end{tabular}     \\
\rowcolor[HTML]{EFEFEF}$\mathbf{K}$               &\begin{tabular}[c]{@{}l@{}} Transition\\ kernel\end{tabular}                                  & $\mathbf{k}_1$         &\begin{tabular}[c]{@{}l@{}} $1$-st transitional\\ kernel constraint\end{tabular} \\
\bottomrule
\end{tabular}
}
\end{center}
\label{tab:notation}
\end{table}
\vspace{-3ex}

\subsection{Federated Learning for Autonomous Driving}
We consider each autonomous vehicle as a data silo. Our goal is to collaboratively train a global driving policy $\theta$ from $N$ silos by aggregating all local learnable weights $\theta_i$ of each silo. Each silo computes the current gradient of its local loss function and then updates the model parameter using an optimizer. 
Mathematically, in the local update stage, at each silo $i$, in each iteration $k$, the model weights of the local silo can be computed as:

\begin{equation}
\begin{aligned}
 \theta_i\left(k + 1\right) =
    {\theta}_i\left(k\right)-\alpha_{k}\frac{1}{m}\sum^m_{h=1}\nabla  \mathcal{L}_{\rm {lr}}\left({\theta}_i\left(k\right),\xi_i^h\left(k\right)\right)  
\end{aligned}
\label{eq:intersilo}
\end{equation}
where $\mathcal{L}_{\rm {lr}}$ is the local regression loss for autonomous steering. To update the global model, each silo interacts with the associated ones through a predefined topology: 

\begin{equation}
\begin{aligned}
 \theta\left(k + 1\right) =
    \sum^{N-1}_{i = 0 }\vartheta_i{\theta}_i\left(k\right)
\end{aligned}
\label{eq:global_theta}
\end{equation}

In practice, the local model in each silo is a deep network that takes the RGB images as inputs and predicts the associated steering angles~\cite{nguyen2022deep}.

\section{Federated Autonomous Driving with Contrastive Divergence Loss}
\subsection{Overview}
\textbf{Motivation.} Due to the non-IID problem, the federated learning algorithms for autonomous driving only achieve good results when two conditions are met: \textit{i)} the local silo $i$ can effectively learn from its local data, and \textit{ii)} the synchronization between nearby silos is sufficient to minimize the effect of the non-IID problem. Currently, most recent works address both of these problems by optimizing the accumulation progress and optimizer~\cite{li2022feel,liang2022federated,parekh2023gefl}, proposing new topologies~\cite{he2021bift}, or utilizing high-performed deep networks that are robust to the non-identical characteristic of distributed data~\cite{bouzinis2021wireless, zhang2021end}.
However, according to~\cite{duan2020self}, naively adopting high-performed deep architecture on centralized local learning and its corresponding optimizations into federated scenarios can increase the weight variance of local silo weights during the weights accumulation between silos. As a consequence, it affects the convergence ability of the model and may also cause divergence.

\begin{figure}[ht]
  \centering
  \includegraphics[width=0.98\linewidth]{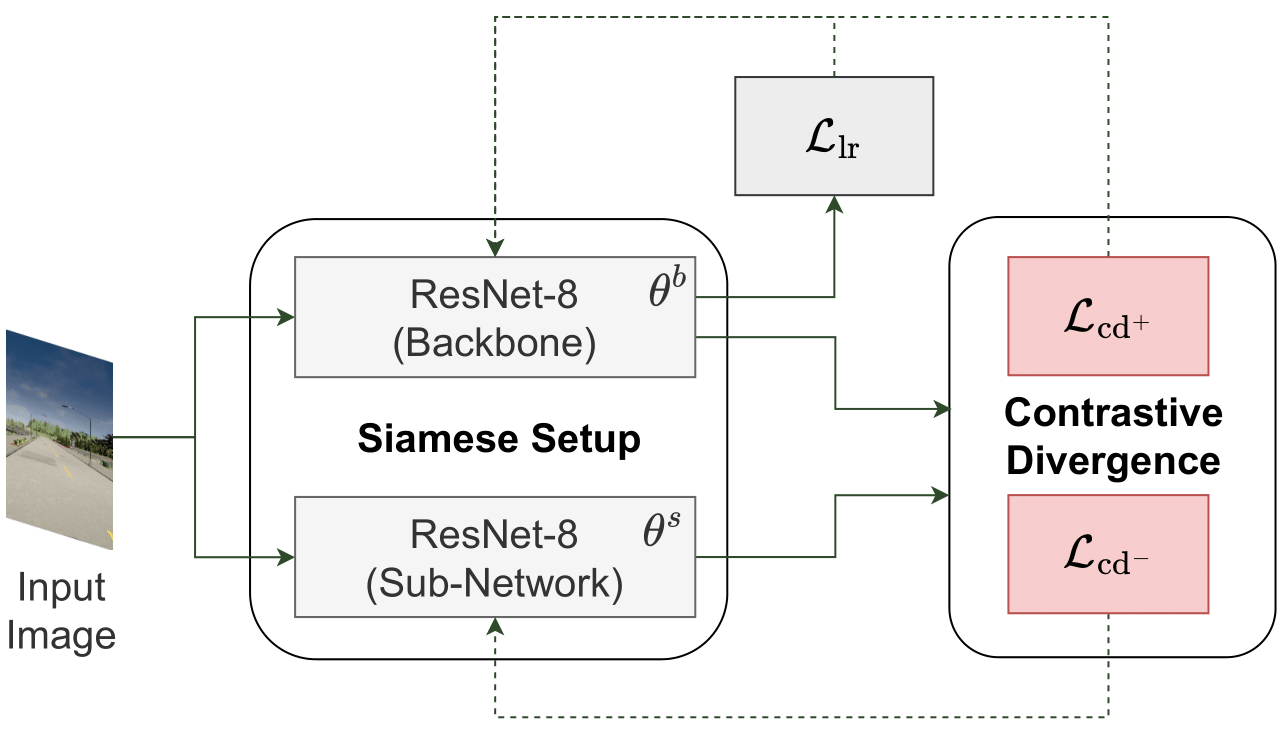}
 \caption{The Siamese setup when our CDL is applied for training federated autonomous driving model. ResNet-8 is used in the backbone and sub-network in the Siamese setup. During inference, the sub-network will be removed. Dotted lines represent the backward process. Our
 CDL has two components: the positive contrastive divergence loss $\mathcal{L}_{\rm {cd^+}}$ and the negative regularize term $\mathcal{L}_{\rm {cd^-}}$. The local regression loss $\mathcal{L}_{\rm {lr}}$ for automatic steering prediction is calculated only from the backbone network.
 }
 \label{fig:network}
\end{figure}

\textbf{Siamese Network.} In this work, we propose to address the non-IID problem directly from each local silo by solving the learning good feature problem and synchronization challenge separately. Consequently, we want to have \textit{two networks in each silo}. One network is responsible for learning meaningful features from the local image data, and the other network is responsible for minimizing the distribution gap between the current model weight and other nearby weights. To this end, the Siamese Network~\cite{koch2015siamese} which includes two networks is well-fitted to our needs. In particular, the Siamese network has two branches. 
We consider the first branch (\textit{backbone} network) of the Siamese to have the local regression loss $\mathcal{L}_{\rm {lr}}$ to learn the local image features for autonomous steering and the positive contrastive divergence loss $\mathcal{L}_{\rm {cd^+}}$ to learn knowledge transmitted from neighbor silos. The second branch (\textit{sub-network}) is ultilized to control the divergence factors from knowledge of the backbone through the contrastive regularizer term $\mathcal{L}_{\rm {cd^-}}$ (See Fig.~\ref{fig:network}). 

In practice, the sub-network shares the same weights with the backbone during the first communication round. However, from the next communication round, after the backbone is accumulated using Equation~\ref{eq:intersilo}, the local model of each silo is trained by the contrastive divergence loss. 
The sub-network outputs support features that have the same size as the output features of the backbone. During the training, we expect that the weights between the backbone and the sub-network should not have significant differences while applying the contrastive divergence loss. The weights of all silos are synchronized whenever gradients from the learning process of the backbone and the sub-network of all silos are not significantly different.

\subsection{Contrastive Divergence Loss}
\label{subsec:CDL}
In practice, we observe that the early stages of federated learning mostly have poor accumulated models. Different from other works that deal with the non-IID problem by optimizing the accumulation step whenever silos transmit their models, we directly decrease the effect of divergence factors during the local learning process of each silo. To achieve that, we reduce the distance between distribution from accumulated weights $\theta^b_i$ at silo $i$ in the backbone network, which contains information from other silos known as divergence factors, and its $i$-th silo weights $\theta^s_i$ in the sub-network, which only contains knowledge learned from local data. When the distribution between silos has been synchronized at an acceptable rate, we lower the effectiveness of the sub-network and focus more on the steering angle prediction task. Our proposed Contrastive Divergence Loss is motivated by the contrastive loss from the original Siamese Network~\cite{koch2015siamese} and is defined as:

\begin{equation}
\begin{aligned}
\mathcal{L}_{\rm {cd}} &= \beta \mathcal{L}_{\rm {cd^+}} + (1-\beta) \mathcal{L}_{\rm {cd^-}} \\
&=
\beta \mathcal{H}(\theta^b_i, \theta^s_i) + (1-\beta) \mathcal{H}(\theta^s_i,\theta^b_i)
\end{aligned}
\label{eq:positive_contrastive_loss}
\end{equation}
where $\mathcal{L}_{\rm {cd^+}}$ is the positive contrastive divergence term and $\mathcal{L}_{\rm {cd^-}}$ is the negative regularizer term; $\mathcal{H}$ is the Kullback-Leibler Divergence loss function~\cite{joyce2011kullback}:

\begin{equation}
\mathcal{H}(\hat{y},y) = \sum \mathbf{f}(\hat{y}) \log\left(\frac{\mathbf{f}(\hat{y})}{\mathbf{f}(y)}\right)
\label{eq:KLDiv}
\end{equation}
where $\hat{y}$ is the predicted representation, $y$ is dynamic soft label. 

Consider $\mathcal{L}_{\rm {cd^+}}$ in Equation~\ref{eq:positive_contrastive_loss} as a Bayesian statistical inference task, 
our goal is to estimate the model parameters $\theta^{b*}$ by minimizing the Kullback-Leibler divergence $\mathcal{H}(\theta^b_i, \theta^s_i)$ between the measured regression probability distribution of the observed local silo $P_0 (x|\theta^s_i)$ and the accumulated model $P (x|\theta^b_i)$. Hence, we can assume that the model distribution has a form of $P (x|\theta^b_i) = e^{-E(x,\theta^b_i)}/Z(\theta^b_i)$, where $Z(\theta^b_i)$ is the normalization term. However, evaluating the normalization term $Z(\theta^b_i)$ is not trivial, which leads to risks of getting stuck in a local minimum. Inspired by Hinton~\cite{hinton2002training}, we use samples obtained through a Markov Chain Monte Carlo (MCMC) procedure with a specific initialization strategy to deal with the mentioned problem. Additionally inferred from Equation~\ref{eq:intersilo}, the $\mathcal{L}_{\rm {cd^+}}$ can be expressed under the SGD algorithm in a local silo by setting:
\begin{equation}
\small
\begin{aligned}
\mathcal{L}_{\rm {cd^+}} = -\sum_{x}P_0 (x|\theta^s_i)\frac{\partial E(x;\theta^b_i)}{\partial \theta^b_i} + \sum_{x}Q_{\theta^b_i} (x|\theta^s_i)\frac{\partial E(x;\theta^b_i)}{\partial \theta^b_i} 
\end{aligned}
\label{eq:CD}
\end{equation}
where $Q_{\theta^b_i} (x|\theta^s_i)$ is the measured probability distribution on the samples obtained by initializing the chain at $P_0 (x|\theta^s_i)$ and running the Markov chain forward for a defined step. 

Consider $\mathcal{L}_{\rm {cd^-}}$ regularizer in Equation~\ref{eq:positive_contrastive_loss} as a Bayesian statistical inference task, we can calculate $\mathcal{L}_{\rm {cd^-}}$  as in Equation~\ref{eq:CD}, however, the role of $\theta^s$ and $\theta^b$ is inverse:

\begin{equation}
\small
\begin{aligned}
\mathcal{L}_{\rm {cd^-}}=-\sum_{x}P_0 (x|\theta^b_i)\frac{\partial E(x;\theta^s_i)}{\partial \theta^s_i} + \sum_{x}Q_{\theta^s_i} (x|\theta^b_i)\frac{\partial E(x;\theta^s_i)}{\partial \theta^s_i}
\end{aligned}
\label{eq:CD_expanded}
\end{equation}

We note that although Equation~\ref{eq:CD} and Equation~\ref{eq:CD_expanded} share the same structure, the key difference is that while the weight $\theta^b_i$ of the backbone is updated by the accumulation process from Equation~\ref{eq:global_theta}, the weight $\theta^s_i$ of the sub-network, instead, is not. This lead to different convergence behavior of contrastive divergence in $\mathcal{L}_{\rm {cd^+}}$ and $\mathcal{L}_{\rm {cd^-}}$. The negative regularizer term $\mathcal{L}_{\rm {cd^-}}$ will converge to state $\theta^{s*}_i$ provided $\frac{\partial E}{\partial \theta^s_i}$ is bounded: 
\begin{equation}
g(x,\theta^s_i) = \frac{\partial E(x;\theta^s_i)}{\partial \theta^s_i} -\sum_{x}P_0 (x|(\theta^b_i,\theta^s_i))\frac{\partial E(x;\theta^s_i)}{\partial \theta^s_i} 
\end{equation}
and
\begin{equation}
\footnotesize
\begin{aligned}
(\theta^s_i - \theta^{s*}_i)\cdot\left\{ \sum_{x}{P_0(x)g(x,\theta^s_i)} - \sum_{x',x}{P_0(x')\mathbf{K}^m_{\theta^s_i}(x',x)g(x,\theta^{s*}_i})\right\}\geq \\
\mathbf{k}_1|\theta^s_i-\theta^{s*}_i|^2
\end{aligned}
\end{equation}
for any $\mathbf{k}_1$ constraint. Note that, $\mathbf{K}^m_{\theta^s}$ is the transition kernel. The proof for the above result is analyzed in~\cite{yuille2004convergence}.


Note that the negative regularizer term $\mathcal{L}_{\rm {cd^-}}$ is only used in training models on local silos. Thus, it does not contribute to the accumulation process of federated training. 



\subsection{Total Training Loss}
\textbf{Local Regression Loss.}
We use mean square error (MAE)  to compute loss for predicting the steering angle in each local silo. Note that, we only use features from the backbone for predicting steering angles.
\begin{equation}
\mathcal{L}_{\rm {lr}} = \text{MAE}(\theta^b_i, \hat{\xi}_i  )
\label{eq:MAE}
\end{equation}
where $\hat{\xi}_i$  is the ground-truth steering angle of the data sample $\xi_i$ collected from silo $i$. 

\textbf{Local Silo Loss.} The local silo loss computed in each communication round at each silo before applying the accumulation process is described as:
\begin{equation}
\mathcal{L}_{{\rm{final}}} = \mathcal{L}_{\rm {lr}} + \mathcal{L}_{\rm {cd}}
\label{eq:final_local}
\end{equation}

In practice, we observe that both the contrastive divergence loss $\mathcal{L}_{\rm {cd}}$ to handle the non-IID problem and the local regression loss $\mathcal{L}_{\rm {lr}}$ for predicting the steering angle is equally important and indispensable. Hence, we do not set a parameter to control their contributions in Equation~\ref{eq:final_local}.




Combining all losses together, at each iteration $k$, the update in the backbone network is defined as: 
\begin{equation}
\begin{aligned}
 &\theta^b_i\left(k + 1\right) \\
 & =\begin{cases}
    \sum_{j \in \mathcal{N}_i^{+} \cup{\{i\}}}\textbf{A}_{i,j}{\theta}^b_{j}\left(k\right), \textit{ \quad \quad \quad if k} \equiv 0 \pmod{u + 1},\\
    {\theta}^b_i\left(k\right)-\alpha_{k}\frac{1}{m}\sum^m_{h=1}\nabla \mathcal{L}_{\rm {b}}\left({\theta}^b_i\left(k\right),\xi_i^h\left(k\right)\right),  \text{otherwise.}
\end{cases}
\end{aligned}
\label{eq:ori_backbone_DFL}
\end{equation}
where $\mathcal{L}_{\rm {b}} = \mathcal{L}_{\rm {lr}} + \mathcal{L}_{\rm {cd^+}}$, $u$ is the number of local updates.

In parallel, the update in the sub-network at each iteration $k$ is described as:
\begin{equation}
\begin{aligned}
 &\theta^s_i\left(k + 1\right) ={\theta}^s_i\left(k\right)-\alpha_{k}\frac{1}{m}\sum^m_{h=1}\nabla \mathcal{L}_{\rm {cd^-}}\left({\theta}^s_i\left(k\right),\xi_i^h\left(k\right)\right)
\end{aligned}
\label{eq:ori_support_DFL}
\end{equation}

\section{Experiment}

\subsection{Implementation}

\textbf{Dataset.} We use three datasets (Table~\ref{tab:datasets}) in our experiment: Udacity+~\cite{udacity2016data}, Gazebo Indoor~\cite{nguyen2022deep}, and Carla Outdoor dataset~\cite{nguyen2022deep}. Gazebo and Calar are non-IID datasets while Udacity+ is the non-IID version of the Udacity dataset.  

\begin{table}[!h]
\caption{The Statistic of Datasets in Our Experiments.
}
\begin{center}
\small
\setlength{\tabcolsep}{0.5 em} 
{\renewcommand{\arraystretch}{1.2}
\begin{tabular}{c|c|c|c|c}
\hline
\multirow{3}{*}{\textbf{Dataset}} & \multirow{3}{*}{\textbf{\begin{tabular}[c]{@{}c@{}}Total \\ samples\end{tabular}}} & \multicolumn{3}{c}{\textbf{\begin{tabular}[c]{@{}c@{}}Average samples  in each silo\end{tabular}}}                                                                            \\ \cline{3-5} 
                                  &                                                                                    & \multirow{2}{*}{\textit{\begin{tabular}[c]{@{}c@{}}Gaia~\cite{knight2011internetzoo}\\ (11 silos)\end{tabular}}} & \multirow{2}{*}{\textit{\begin{tabular}[c]{@{}c@{}}NWS~\cite{awscloud}\\ (22 silos)\end{tabular}}}& \multirow{2}{*}{\textit{\begin{tabular}[c]{@{}c@{}}Exodus~\cite{knight2011internetzoo}\\ (79 silos)\end{tabular}}} \\
                                  & & &  & \\ \hline
Udacity+  & 38,586 &3,508  &1,754  &488\\ \hline
Gazebo  &66,806 &6,073 &3,037 &846\\ 
\hline
Carla   &73,235 &6,658 &3,329 &927\\ \hline
\end{tabular}
}
\end{center}

\label{tab:datasets}
\end{table}

\begin{table*}[ht]
\caption{Performance comparison between different methods. The Gaia topology is used.
}
\begin{center}
\small
\setlength{\tabcolsep}{0.39 em} 
{\renewcommand{\arraystretch}{1.1}
\begin{tabular}{l|c|c|c|c|c|c|c|c|c|c}
\hline
\multirow{2}{*}{\textbf{Model}} &
\multirow{2}{*}{\textbf{\begin{tabular}[c]{@{}c@{}}Main\\ Focus\end{tabular}}} &
\multirow{2}{*}{\textbf{\begin{tabular}[c]{@{}c@{}}Learning\\ Method\end{tabular}}} & \multicolumn{3}{c|}{\textbf{RMSE}} & \multicolumn{3}{c|}{\textbf{MAE}} & \multirow{2}{*}{\textbf{\begin{tabular}[c]{@{}c@{}}\# Training \\ Parameters\end{tabular}}} & \multirow{2}{*}{\textbf{\begin{tabular}[c]{@{}c@{}}Avg. Cycle \\ Time (ms)\end{tabular}}} \\ \cline{4-9}
 &  & & {\textit{\textbf{Udacity+}}} & {\textit{\textbf{Gazebo}}} & \textit{\textbf{Carla}} & {\textit{\textbf{Udacity+}}} & {\textit{\textbf{Gazebo}}} & \textit{\textbf{Carla}} &  &  \\ \hline
Random~\cite{loquercio2018dronet} &\_ & \multirow{2}{*}{\_} & {0.358} & {0.117} & 0.464 & {0.265} & {0.087} & 0.361 & \_ &\_  \\ 
Constant~\cite{loquercio2018dronet} & Statistical & & {0.311} & {0.092} & 0.348 & {0.209} & {0.067} & 0.232 & \_ & \_ \\ \hline
\hline
Inception~\cite{szegedy2016rethinking} &\multirow{4}{*}{\begin{tabular}[c]{@{}c@{}}Architecture\\Design\end{tabular}} & \multirow{4}{*}{CLL~\cite{loquercio2018dronet}} & 0.209 & {0.085} & 0.297 &0.197  & {0.062} & 0.207 & 21,787,617 &\_  \\ 
MobileNet~\cite{sandler2018mobilenetv2} & & &0.193  & {0.083} & 0.286 & 0.176 & {0.057} & 0.200 & 2,225,153 & \_ \\ 
VGG-16~\cite{simonyan2014very} & & &0.190  & {0.083} & 0.316 &0.161  & {0.050} & 0.184 & 7,501,587 &\_  \\ 
DroNet~\cite{loquercio2018dronet} & & & {0.183} & {0.082} & 0.333 & {0.150} & {0.053} & 0.218 & 314,657 & \_ \\ \hline
\hline
FedAvg~\cite{mcmahan2017communication} &\multirow{3}{*}{\begin{tabular}[c]{@{}c@{}}Aggregation\\Optimization\end{tabular}} & \multirow{3}{*}{SFL~\cite{sattler2019robust}}  &0.212  &0.094  &0.269  &0.185  &0.064  &0.222  & 314,657 & 152.4  \\
FedProx~\cite{li2018federated} & & &0.152  &0.077  &0.226   & 0.118 &0.063  &0.151  & 314,657 & 111.5 \\
STAR ~\cite{sattler2019robust} & & &0.179  &0.062  &0.208  &0.149  &0.053  & 0.155 & 314,657 &299.9  \\ \hline 
\hline
MATCHA~\cite{wang2019matcha} &\multirow{3}{*}{\begin{tabular}[c]{@{}c@{}}Topology\\Design\end{tabular}} & \multirow{3}{*}{DFL~\cite{nguyen2022deep}} &0.182  & 0.069 & 0.208  &0.148  &0.058  &0.215  & 314,657 &171.3  \\ 
MBST~\cite{prim1957shortest,marfoq2020throughput} & & &0.183   &0.072  &0.214  &0.149  &0.058  &0.206  & 314,657 & 82.1 \\ 
FADNet~\cite{nguyen2022deep} & & & {0.162} & {0.069} & 0.203 & {0.134} & {0.055} & 0.197 & 317,729 & \textbf{62.6}  \\ \hline 
\hline
\multirow{3}{*}{\textbf{CDL (ours)}} & \multirow{3}{*}{\begin{tabular}[c]{@{}c@{}}Loss\\Optimization\end{tabular}} & CLL~\cite{nguyen2022deep} & 0.169 & 0.074 & 0.266 &0.149  &0.053  &0.172  & {629,314} &\_ \\
& & SFL~\cite{nguyen2022deep} &0.150 & \textbf{0.060} & 0.208 &0.104  &\textbf{0.052}  &0.150  & {629,314} & 102.2  \\
& &  DFL~\cite{nguyen2022deep} & \textbf{0.141} & 0.062 & \textbf{0.183} & \textbf{0.083} & \textbf{0.052} & \textbf{0.147} & 629,314 &72.7  \\\hline
\end{tabular}
}
\end{center}

\label{tab:sota}
\end{table*}

\textbf{Network Topology.} Following~\cite{marfoq2020throughput}, we conduct experiments on three federated topologies: the Internet Topology Zoo~\cite{knight2011internetzoo} (Gaia), the North American data centers~\cite{awscloud} (NWS), and the Zoo Exodus network (Exodus)~\cite{knight2011internetzoo}. We use Gaia topology in our main experiment and provide the comparison of two other topologies in our ablation study. 

\textbf{Training.} The model in a silo is trained with a batch size of $32$ and a learning rate of $0.001$ using Adam optimizer. In each communication round, the local training process is done in each silo before their models are transmitted and aggregated using Equation~\ref{eq:global_theta}. The training process is conducted with $3,600$ communication rounds. We apply the simulation as in~\cite{nguyen2022deep} for training with an NVIDIA 1080 GPU. 

\textbf{Baselines.}
We compare our results with various recent methods in different learning scenarios, including Random baseline and the Constant baseline~\cite{loquercio2018dronet}. For the Centralized Local Learning (CLL) scenario, Inception-V3~\cite{szegedy2016rethinking}, MobileNet-V2~\cite{sandler2018mobilenetv2}, VGG-16~\cite{simonyan2014very}, and Dronet~\cite{loquercio2018dronet} are used as baselines. For the Server-based Federated Learning (SFL) scenario, we compare our method with FedAvg~\cite{mcmahan2017communication}, FedProx~\cite{li2018federated}, and STAR~\cite{sattler2019robust}. For the Decentralized Federated Learning (DFL), MATCHA~\cite{wang2019matcha}, MBST~\cite{marfoq2020throughput}, and FADNet~\cite{nguyen2022deep} are used as baselines.
We use the Root Mean Square Error (RMSE) and Mean Absolute Error (MAE) to evaluate the effectiveness of models. Besides, wall-clock time (ms) is used to calculate the time needed for training each method.

\subsection{Qualitative Results}
Table~\ref{tab:sota} summarises the performance of our method and recent state-of-the-art approaches. This table clearly shows the proposed CDL under Siamese setup with two ResNet-8 outperforms other methods by a large margin. In particular, our proposal significantly reduces both RMSE and MAE in all three datasets, including Udacity+, Carla, and Gazebo. 
As a loss function, CDL does not increase the number of parameters of the network. However, under the Siamese setup, the model size during the training is increased as the Siamese requires an additional sub-network. Moreover, we can see that our CDL with ResNet-8 outperforms other baselines by a large margin in the DFL learning scenario, and by a moderate value in  SFL and CLL setup. 

\subsection{Contrastive Divergence Loss Analysis}
\label{eff_def}


\begin{table}[]
\caption{Performance under different topologies.}
\begin{center}
\resizebox{\linewidth*41/40}{!}{
\setlength{\tabcolsep}{0.25 em} 
{\renewcommand{\arraystretch}{1.2}
\begin{tabular}{c|c|c|c|c}
\hline
\multirow{2}{*}{\textbf{Topology}}                                          & \multirow{2}{*}{\textbf{Architecture}} & \multicolumn{3}{c}{\textbf{Dataset}}                                     \\ \cline{3-5} 
                                                                           &                                        & \textit{\textbf{Udacity+}} & \textit{\textbf{Gazebo}} & \textit{\textbf{Carla}} \\ \hline
\multirow{3}{*}{\begin{tabular}[c]{@{}c@{}}Gaia\\ (11 silos)\end{tabular}} & DroNet~\cite{loquercio2018dronet}                                 & 0.177\improvecolor($\downarrow$0.036) & 0.073\improvecolor($\downarrow$0.011) & 0.244\improvecolor($\downarrow$0.061)            \\ 
                                                                           & FADNet~\cite{nguyen2022deep}                                  & 0.162\improvecolor($\downarrow$0.021) & 0.069\improvecolor($\downarrow$0.007) & 0.203\improvecolor($\downarrow$0.020)     \\
                                                                           & \textbf{CDL (ours)}                                  & \textbf{0.141} & \textbf{0.062} & \textbf{0.183}         \\ \hline
\multirow{3}{*}{\begin{tabular}[c]{@{}c@{}}NWS\\ (22 silos)\end{tabular}}  & DroNet~\cite{loquercio2018dronet}                                 & 0.183\improvecolor($\downarrow$0.045)                    & 0.075\improvecolor($\downarrow$0.017)                & 0.239\improvecolor($\downarrow$0.057)            \\ 
                                                                           & FADNet~\cite{nguyen2022deep}                                   &0.165\improvecolor($\downarrow$0.027)                     & 0.070\improvecolor($\downarrow$0.012)                 & 0.200\improvecolor($\downarrow$0.018)            \\
                                                                           & \textbf{CDL (ours)}                                  & \textbf{0.138}                     & \textbf{0.058}                 & \textbf{0.182}                 \\ \hline
\multirow{3}{*}{\begin{tabular}[c]{@{}c@{}}Exodus\\ (79 silos)\end{tabular}}  & DroNet~\cite{loquercio2018dronet}                                 & 0.448\improvecolor($\downarrow$0.310)                   & 0.208\improvecolor($\downarrow$0.147)               & 0.556\improvecolor($\downarrow$0.380)         \\
                                                                           & FADNet~\cite{nguyen2022deep}                                   & 0.179\improvecolor($\downarrow$0.041)                    &0.081\improvecolor($\downarrow$0.02)                 &0.238\improvecolor($\downarrow$0.062)             \\ 
                                                                           & \textbf{CDL (ours)}                                   & \textbf{0.138}                    & \textbf{0.061}                & \textbf{0.176}               \\ \hline
\end{tabular}
}
}
\end{center}
\label{tab:cross_silo}
\end{table}

\textbf{CDL on Different Topologies.} In practice, it is usually more challenging to train federated algorithms when the topology has more vehicle data silos. To verify the effectiveness of our CDL, we train our method and compare it with other baselines on topologies with different numbers of silos. Table~\ref{tab:cross_silo} illustrates the performance of DroNet, FADNet, and our CDL with ResNet-8 backbone when we train them using DFL under three distributed network infrastructures with different numbers of silos: Gaia (11 silos), NWS (22 silos), and Exodus (79 silos). This table shows that our CDL clearly achieves the highest results in all topology setups, while DroNet meets divergence, and FADNet does not perform well in the Exodus topology which has 79 silos.

\begin{figure}[t]
  \centering
\setlength{\tabcolsep}{0pt}
\begin{tabular}{ccc}
\shortstack{\includegraphics[width=0.49\linewidth, height=0.55\linewidth]{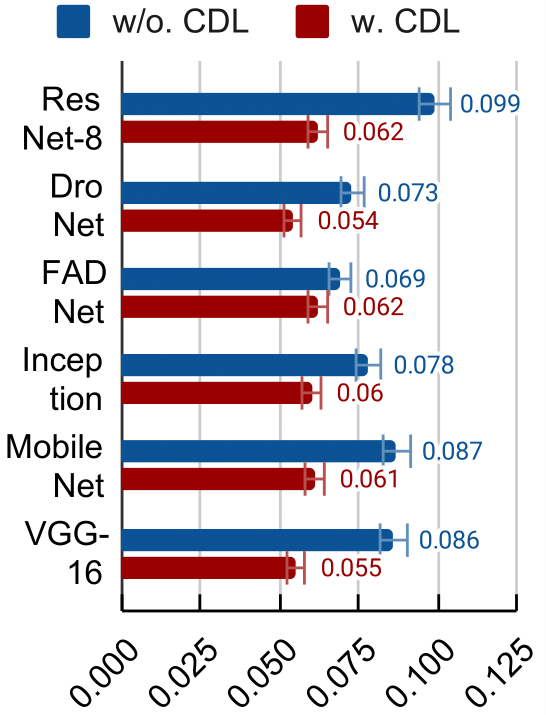}\\\footnotesize (a) Gazebo}&
\shortstack{\includegraphics[width=0.49\linewidth,height=0.55\linewidth]{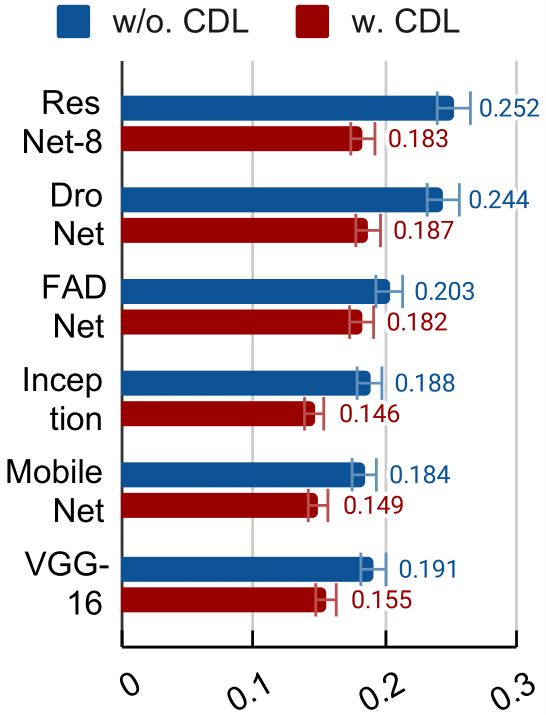}\\ \footnotesize (b) Carla}\\

\end{tabular}
\vspace{-1ex}
    \caption{Performance of CDL under different networks in Siamese setup. 
    }
    \label{fig_fl_topo}
\end{figure}

\begin{figure}[ht]
  \centering
\setlength{\tabcolsep}{0.2 em} 

\begin{tabular}{cc}
\shortstack{\includegraphics[width=0.45\linewidth]{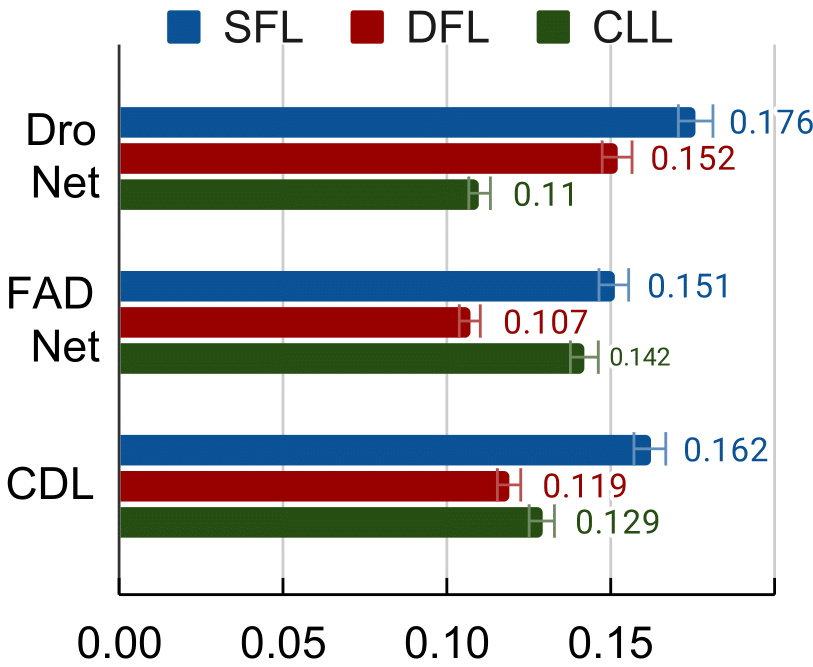}\\ \small (a) IID data: Udacity}& \hspace{1ex}
\shortstack{\includegraphics[width=0.45\linewidth]{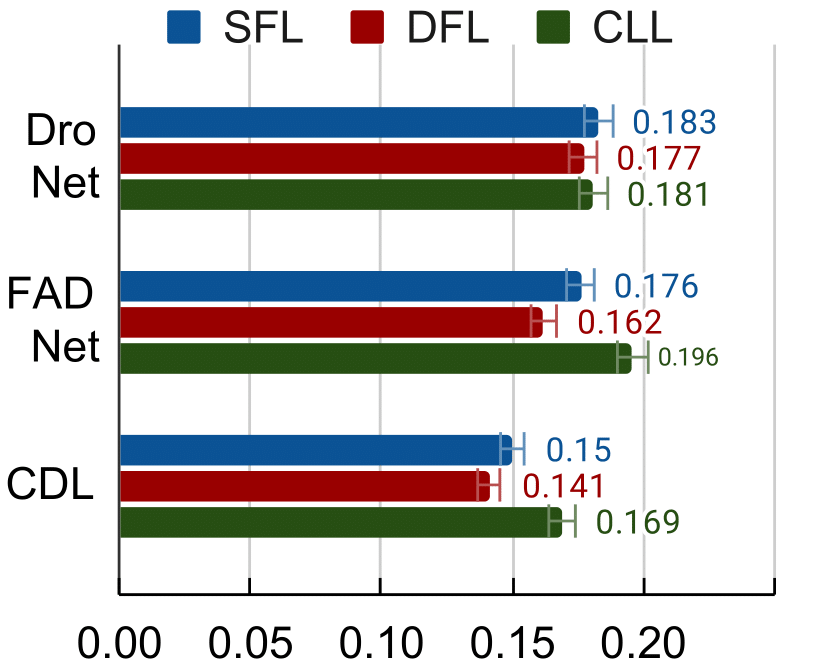}\\ \small (b) Non-IID data: Udacity+}\\
\end{tabular}
    \caption{Performance of different methods on IID dataset (Udacity) and non-IID dataset (Udacity+). 
    }
    \label{fig:IID&NonIID}
\end{figure}

\textbf{CDL on Different Backbones.} Since our proposed CDL is a loss function, it can be applied to different networks under the Siamese setup to improve performance. Figure~\ref{fig_fl_topo} illustrates the effectiveness of CDL when we change the network inside the Siamese to DroNet, FADNet, Inception, MobileNet, and VGG-16 under Gaia Network in the DFL scenario. The results show that CDL works well with different architectures to address the non-IID problem and consistently improving the performance. 

\textbf{CDL on IID Data.} Figure~\ref{fig:IID&NonIID} demonstrates the effectiveness of CDL in different data distributions. Although CDL is designed for dealing with the non-IID problem, it also slightly improves the performance of models trained on IID data distribution. Based on the Siamese setup, CDL inherits the behavior and characteristic of triplet loss form. Since triplet loss is proven to be effective in IID data~\cite{wang2020understanding}, it is clear that CDL can also improve the performance of models when we train them with the IID data.

\subsection{Ablation Study}
\begin{figure}[t]
  \centering
\resizebox{\linewidth}{!}{
\setlength{\tabcolsep}{0pt}
\begin{tabular}{ccc}
 \shortstack{\includegraphics[width=0.33\linewidth]{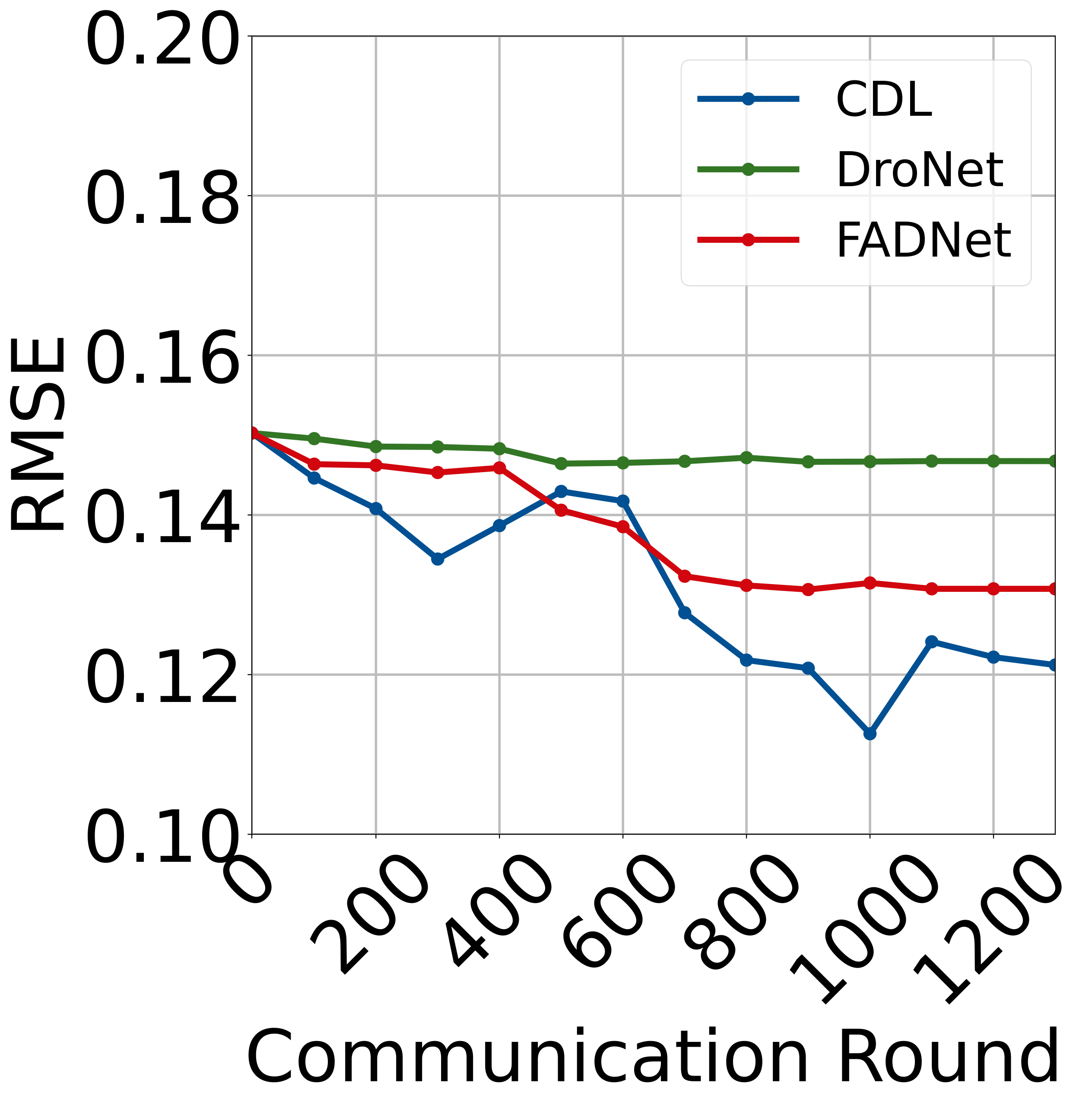}}&
 \shortstack{\includegraphics[width=0.33\linewidth]{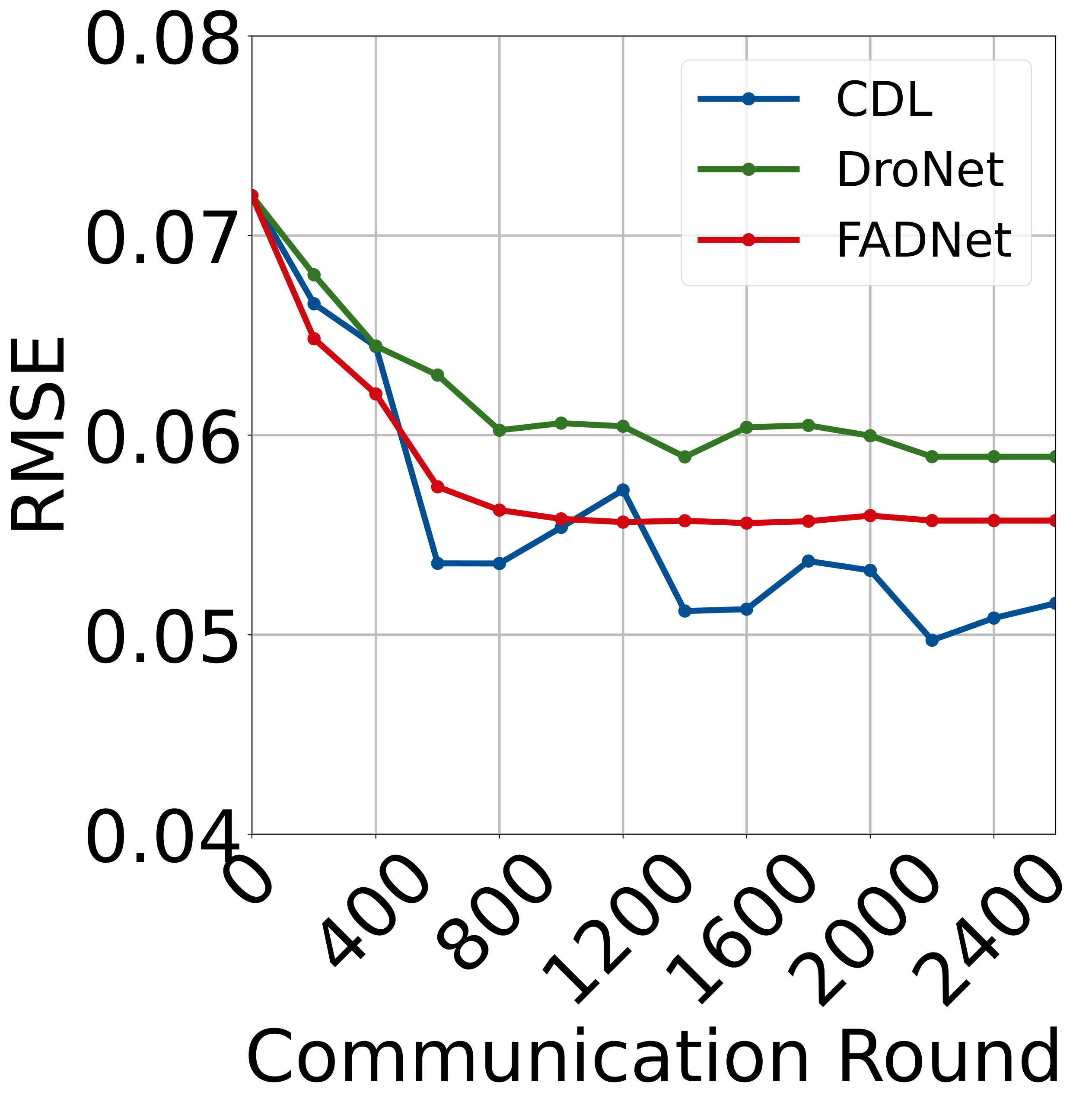}}&
 \shortstack{\includegraphics[width=0.33\linewidth]{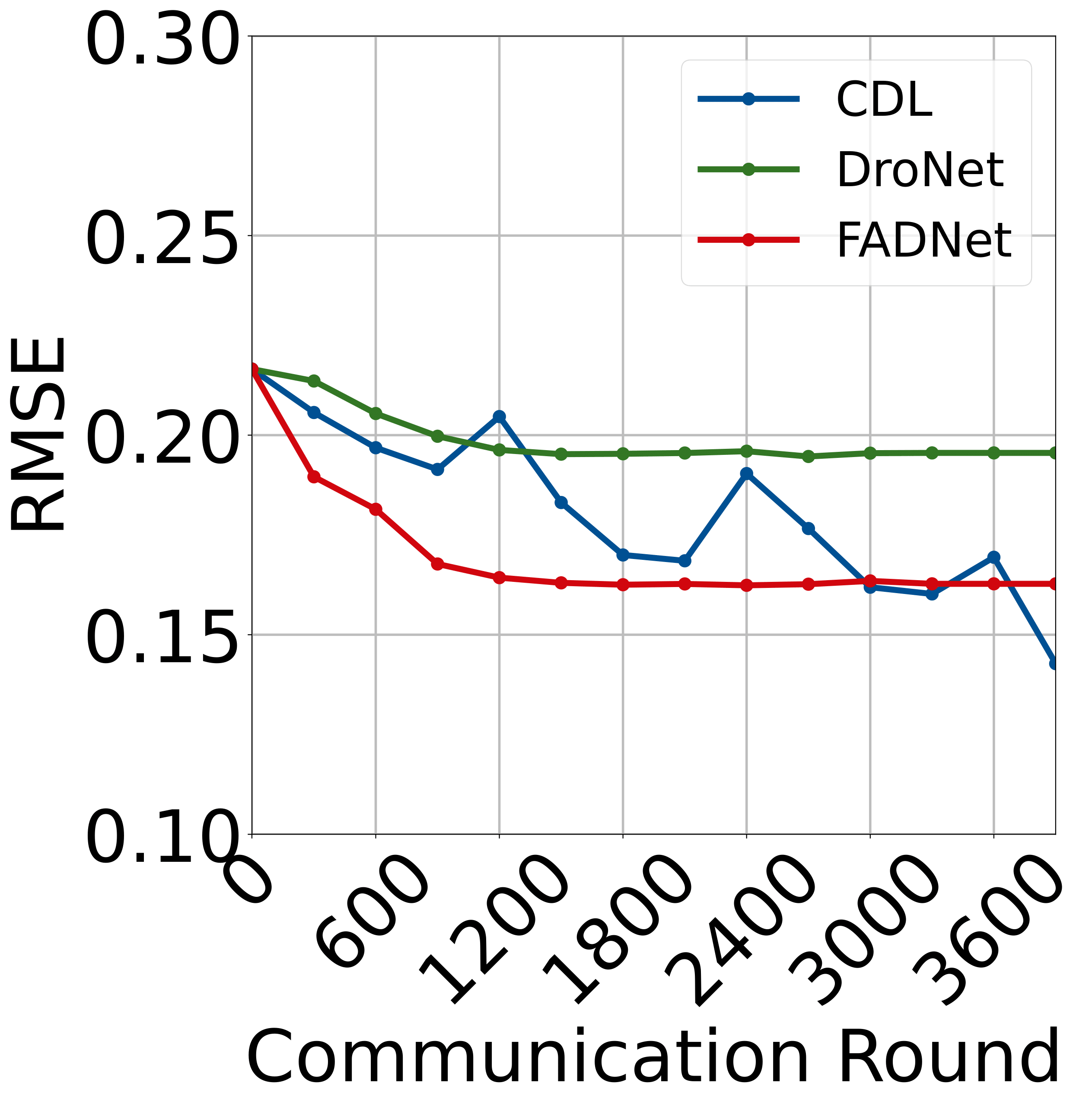}}\\
\shortstack{\includegraphics[width=0.33\linewidth]{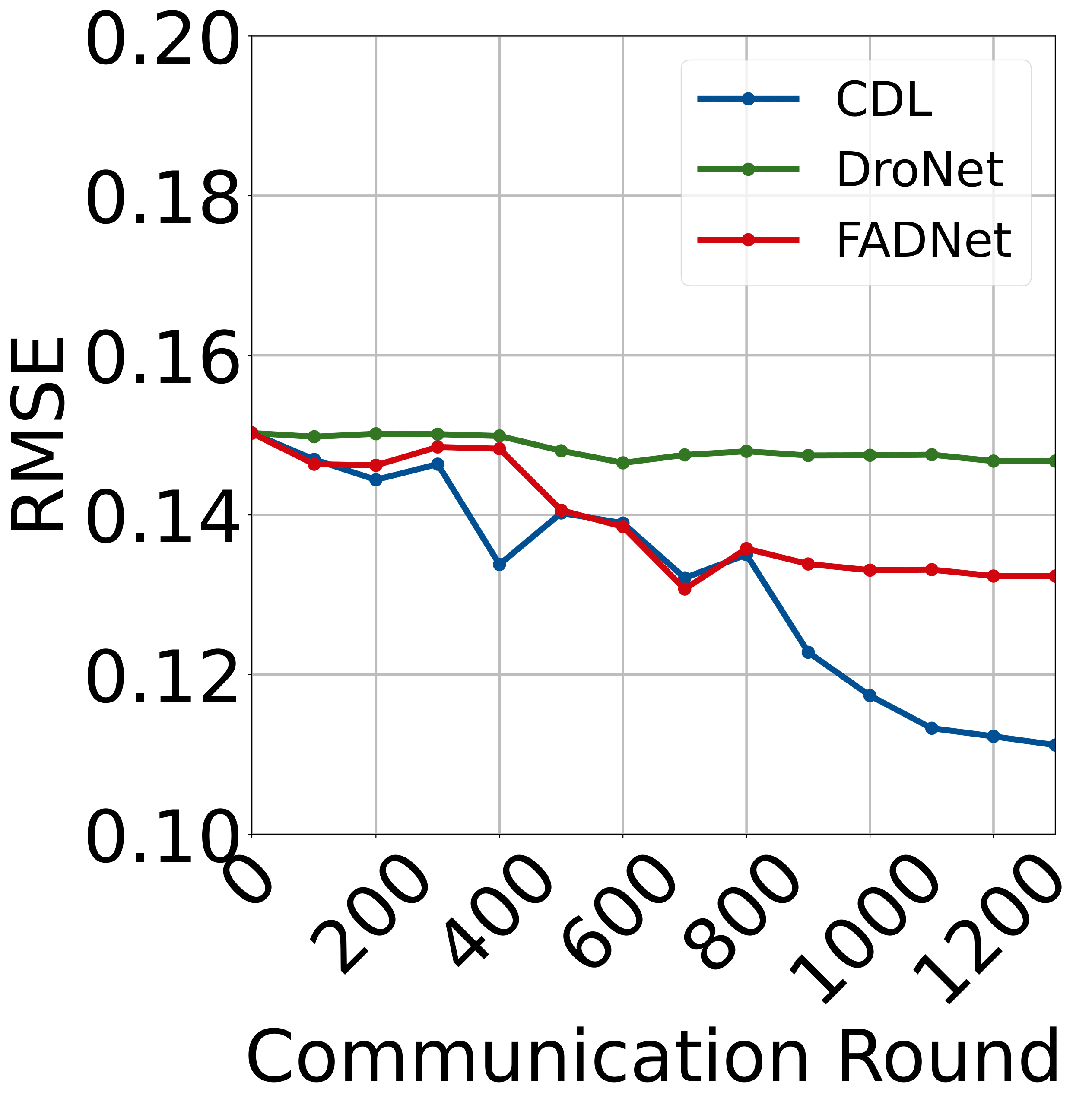}\\\small{(a) Udacity+}}&
\shortstack{\includegraphics[width=0.33\linewidth]{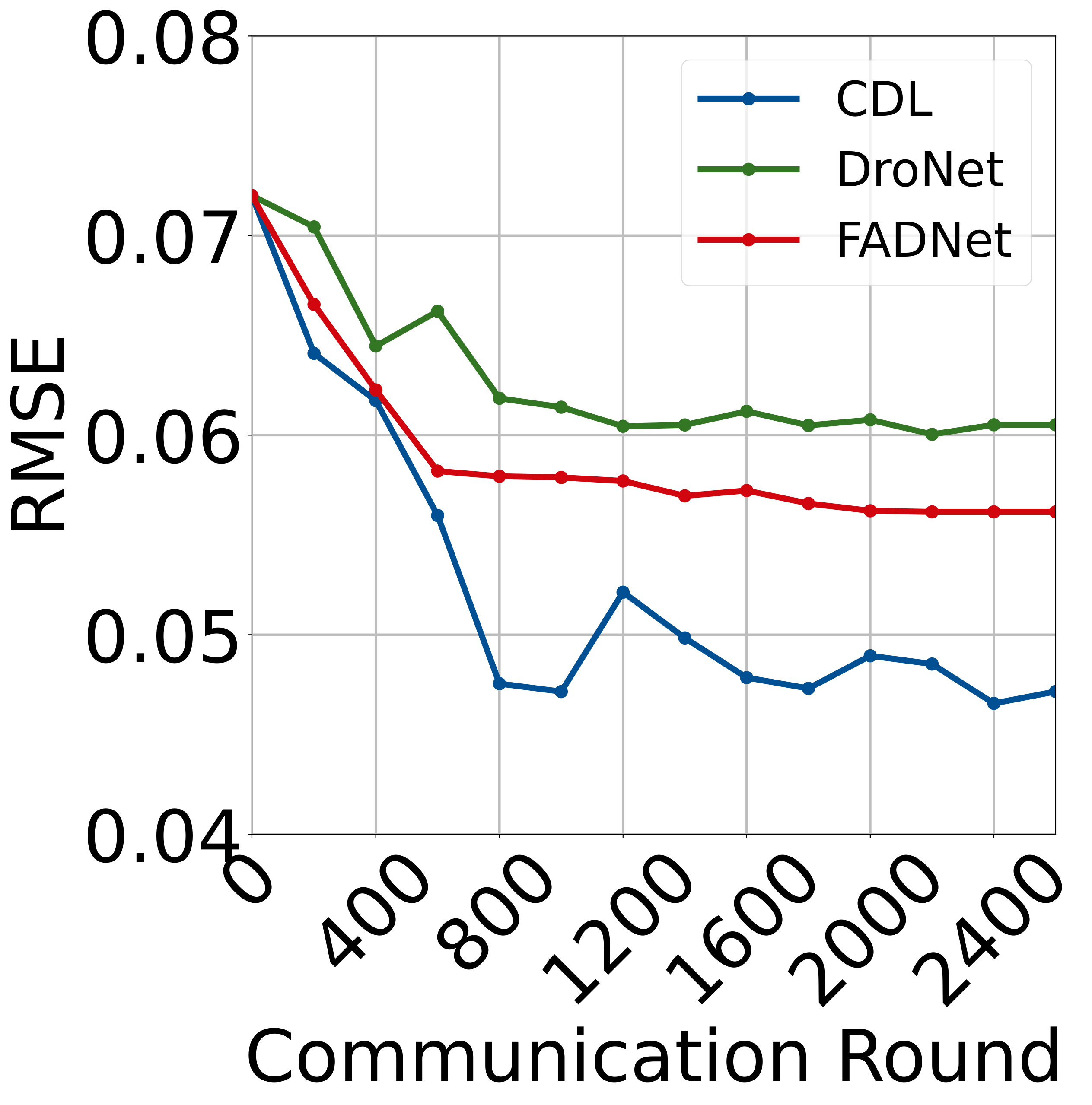}\\\small{(b) Gazebo}}&
\shortstack{\includegraphics[width=0.33\linewidth]{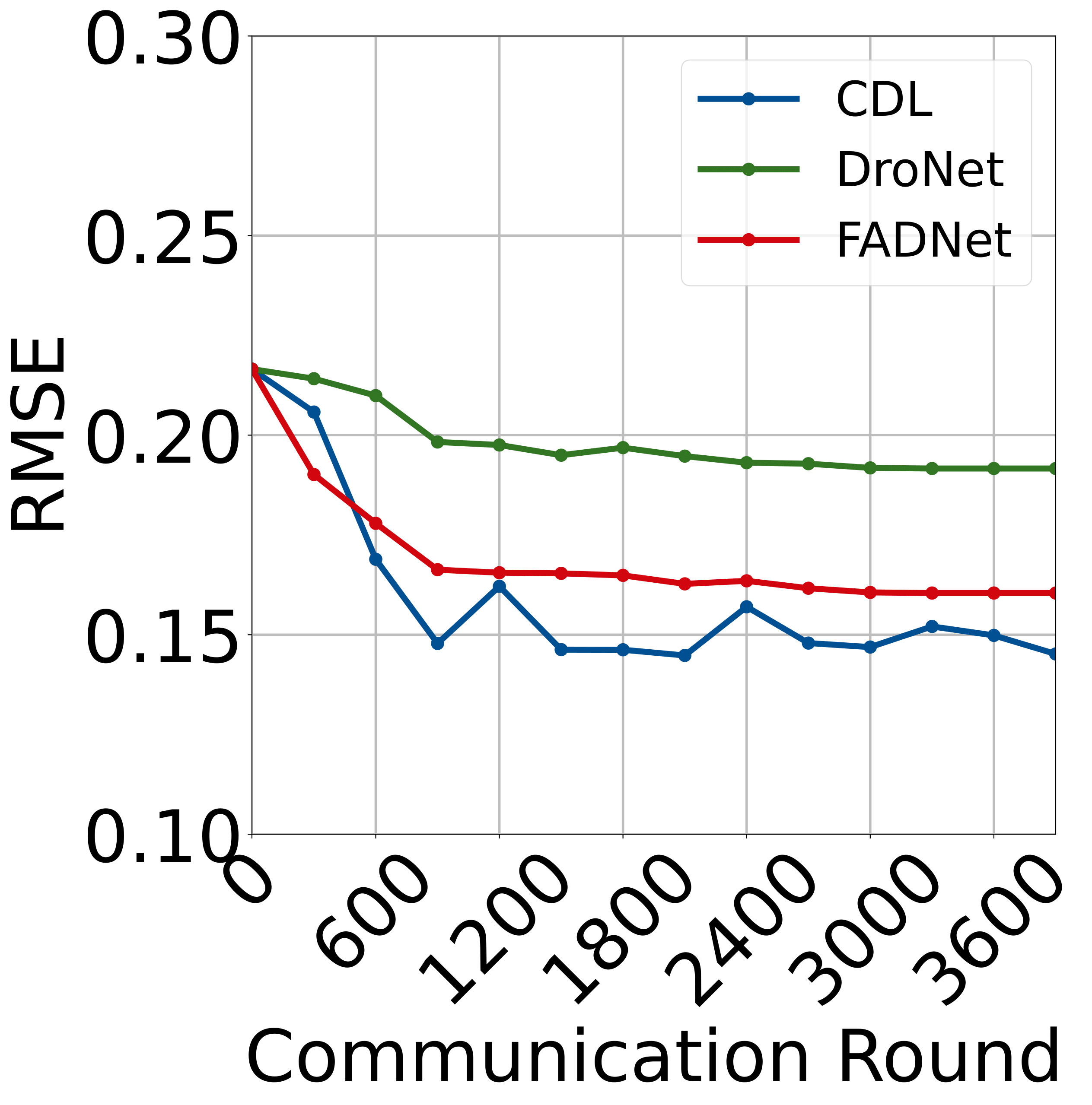}\\\small{(c) Carla}}\\
\end{tabular}
}
    \caption{The convergence ability of different methods under Gaia topology (top row) and
    NWS topology (bottom row).
    }
    \label{fig:convergence}
\end{figure}

\textbf{Convergence Analysis.} Figure~\ref{fig:convergence} illustrates the training results in RMSE of our two baselines DroNet and FADNet as well as our proposed CDL. The results show the convergence ability of mentioned methods in three datasets (Udacity+, Gazebo, Carla) with NWS and Gaia topology. The results indicate that our proposed CDL can reach a better convergence point in comparison with the two baselines. While other methods (Dronet and FADNet) converged with difficulty or do not show good convergence trend, our proposed CDL can get over local optimal points better than other methods and also be less biased into any specific silo.

\textbf{Robotic Demonstration.}
We deploy the global model aggregated from our CDL with ResNet-8 of each vehicle trained on the Non-IID Udacity+ dataset to a standard mobile robot platform. The robot is equipped with an 8-Core ARM v$8.2$ $ 64$-bit CPU and NVIDIA Jetson. Since the inference time of our method is only $42$ ms and the number of parameters is only approximately $630,000$, it can predict steering angles for the robot in real time. More quantitative results can be found in our Demonstration Video.


\subsection{Discussion} Although our method shows promising results, we see several improvement points that can be considered for future work: \textit{i)} Our CDL is designed to mitigate the non-IID problem, thus, the approach's effectiveness heavily relies on the existence of non-IID characteristics within the autonomous driving data. Consequently, in scenarios where the data exhibits relatively consistent distributions across vehicles or lacks significant non-IID factors, the proposed contrastive divergence loss only yields limited performance gains. 
\textit{ii)} Our proposal has been validated on autonomous driving datasets, but not on real vehicles. A real-world study on different driving scenarios involving interactions with pedestrians, cyclists, and other vehicles can further validate our method. And \textit{iii)} While the approach is tailored for autonomous driving, its principles might find applications in other domains where non-IID data is present. Exploring its efficacy in areas such as healthcare, IoT, and industrial settings could broaden its scope of impact.

\section{Conclusion}
We presented a new method to address the non-IID problem in federated autonomous driving using contrastive divergence loss. Our method directly reduces the effect of divergence factors in the learning process of each silo. We analyze the proposed contrastive divergence loss in theories, on various autonomous driving scenarios, under multiple network topologies, and with different FL settings. The experiments on three benchmarking datasets demonstrate that our proposed method performs substantially better than current state-of-the-art approaches. In the future, we plan to test our strategy with more data silos and deploy the trained model using an autonomous vehicle on roads. We will also release our source code to encourage further study.

\bibliographystyle{class/IEEEtran}
\bibliography{class/reference}

\end{document}